\documentclass[pdflatex,sn-mathphys-num]{sn-jnl}

\usepackage{graphicx}%
\usepackage{multirow}%
\usepackage{amsmath,amssymb,amsfonts}%
\usepackage{amsthm}%
\usepackage{mathrsfs}%
\usepackage[title]{appendix}%
\usepackage{xcolor}%
\usepackage{textcomp}%
\usepackage{manyfoot}%
\usepackage{booktabs}%
\usepackage{algorithm}%
\usepackage{algorithmicx}%
\usepackage{algpseudocode}%
\usepackage{listings}%
\usepackage{booktabs}
\usepackage{array}

\theoremstyle{thmstyleone}%
\theoremstyle{thmstyletwo}%

\theoremstyle{thmstylethree}%

\begin{document}

\title[EndoASR]{Development and multi-center evaluation of domain-adapted speech recognition for human-AI teaming in real-world gastrointestinal endoscopy}



\author[1,2,3,4]{\fnm{Ruijie} \sur{Yang}}
\equalcont{These authors contributed equally to this work.}

\author[5,6]{\fnm{Yan} \sur{Zhu}}
\equalcont{These authors contributed equally to this work.}

\author[5,6]{\fnm{Peiyao} \sur{Fu}}
\equalcont{These authors contributed equally to this work.}

\author[3,4]{\fnm{Te} \sur{Luo}}

\author[1,2]{\fnm{Zhihua} \sur{Wang}}

\author[7]{\fnm{Xian} \sur{Yang}}

\author[5,6]{\fnm{Quanlin} \sur{Li}}

\author*[5,6]{\fnm{Pinghong} \sur{Zhou}}\email{zhou.pinghong@zs-hospital.sh.cn}

\author*[3,4]{\fnm{Shuo} \sur{Wang}}\email{shuowang@fudan.edu.cn}

\affil[1]{\orgname{Zhejiang University}, \orgaddress{\city{Hangzhou}, \country{China}}}

\affil[2]{\orgname{Shanghai Institute for Advanced Study, Zhejiang University}, \orgaddress{\city{Shanghai}, \country{China}}}

\affil[3]{\orgname{Shanghai Key Laboratory of MICCAI}, \orgaddress{\city{Shanghai}, \country{China}}}

\affil[4]{\orgname{Digital Medical Research Center, School of Basic Medical Sciences, Fudan University}, \orgaddress{\city{Shanghai}, \country{China}}}

\affil[5]{\orgname{Endoscopy Center and Endoscopy Research Institute, Zhongshan Hospital, Fudan University}, \orgaddress{\city{Shanghai}, \country{China}}}

\affil[6]{\orgname{Shanghai Collaborative Innovation Center of Endoscopy}, \orgaddress{\city{Shanghai}, \country{China}}}

\affil[7]{\orgname{Alliance Manchester Business School, The University of Manchester}, \orgaddress{\city{Manchester}, \country{UK}}}


\abstract{
Automatic speech recognition (ASR) is a critical interface for human–AI interaction in gastrointestinal endoscopy, yet its reliability in real-world clinical settings is limited by domain-specific terminology and complex acoustic conditions. Here, we present EndoASR, a domain-adapted ASR system designed for real-time deployment in endoscopic workflows. 
We develop a two-stage adaptation strategy based on synthetic endoscopy reports, targeting domain-specific language modeling and noise robustness. In retrospective evaluation across six endoscopists, EndoASR substantially improves both transcription accuracy and clinical usability, reducing character error rate (CER) from 20.52\% to 14.14\% and increasing medical term accuracy (Med ACC) from 54.30\% to 87.59\%. 
In a prospective multi-center study spanning five independent endoscopy centers, EndoASR demonstrates consistent generalization under heterogeneous real-world conditions. Compared with the baseline Paraformer model, CER is reduced from 16.20\% to 14.97\%, while Med ACC is improved from 61.63\% to 84.16\%, confirming its robustness in practical deployment scenarios.
Notably, EndoASR achieves a real-time factor (RTF) of 0.005, significantly faster than Whisper-large-v3 (RTF 0.055), while maintaining a compact model size of 220M parameters, enabling efficient edge deployment. Furthermore, integration with large language models demonstrates that improved ASR quality directly enhances downstream structured information extraction and clinician–AI interaction.
These results demonstrate that domain-adapted ASR can serve as a reliable interface for human–AI teaming in gastrointestinal endoscopy, with consistent performance validated across multi-center real-world clinical settings.
}

\keywords{automatic speech recognition, gastrointestinal endoscopy, domain adaptation, human--AI teaming}



\maketitle

\section{Introduction}\label{sec1}
As artificial intelligence (AI) continues to transform healthcare, its transition from controlled research environments to dynamic and unpredictable real-world clinical settings has become a key determinant of whether AI innovations can truly deliver clinical value~\cite{topol2019high, youssef2023external}. In high-acuity procedural environments such as gastrointestinal endoscopy units, this challenge is particularly critical. To be clinically useful, AI systems must demonstrate not only strong algorithmic performance, but also safety, reliability, and seamless integration into routine clinical workflows involving real patients~\cite{kelly2019key}.

Recent advances in medical artificial intelligence are driving a paradigm shift from isolated, task-specific algorithms toward agent-based systems that actively participate in clinical workflows through continuous perception, interaction, and decision-making. In this emerging Human–AI teaming framework, AI systems are no longer passive tools that operate offline, but interactive agents that collaborate with clinicians in real time, adapt to dynamic clinical contexts, and support complex procedural workflows at the point of care. This transition—from algorithm-centric models to embodied clinical AI agents—places increasing emphasis on robust, low-latency perception and natural human–AI interfaces, particularly in time-critical procedural environments \cite{hu2025landscape}.

Within this paradigm, Automatic Speech Recognition (ASR) serves as a foundational interaction interface that enables seamless, hands-free communication between clinicians and AI agents. Spoken language constitutes one of the most natural and efficient modalities for real-time interaction in clinical settings, especially in environments where visual attention and manual operation are fully occupied. In gastrointestinal endoscopy, physicians frequently have both hands engaged with endoscopes and therapeutic instruments, making voice-based interaction uniquely suited for real-time communication, documentation, and AI-assisted support \cite{ng2025evaluating}. Endoscopists routinely verbalize findings, procedural decisions, lesion characteristics, and post-procedural plans during examinations and interventions, and these spoken descriptions directly determine the completeness, accuracy, and timeliness of endoscopy reports.

From a system perspective, ASR represents a critical perception module within agent-based clinical AI architectures, transforming raw acoustic signals into structured linguistic representations that can be further interpreted by large language models and downstream clinical reasoning agents \cite{lee2023machine, blackley2019speech, kumar2024comprehensive}. As large language models become increasingly embedded in clinical decision support, workflow automation, and conversational AI agents, the quality of ASR directly constrains the fidelity of downstream semantic understanding, structured data generation, and human–AI interaction, ultimately influencing patient safety, clinician trust, and system reliability \cite{wiest2025large}. In this sense, ASR functions not merely as a transcription tool, but as a primary user interface for AI–human teaming in procedural medicine.

However, speech perception in endoscopy suites differs fundamentally from everyday conversational speech. Physicians routinely employ highly specialized medical terminology—such as \textit{Clopidogrel}, \textit{Diverticulum}, \textit{Boston Bowel Preparation Scale} (BBPS) scores, Paris classification IIa, and \textit{endoscopic mucosal resection} (EMR)—that is sparsely represented in general-domain training corpora. At the same time, endoscopy rooms are characterized by persistent operational noise from endoscopy towers, flushing and suction devices, alarms, and frequent multi-speaker interactions, resulting in substantial acoustic interference and overlapping speech \cite{schulte2020automatic}. These factors pose significant challenges for real-time, edge-deployed ASR systems operating under strict latency and robustness constraints.

Although state-of-the-art general-purpose ASR models—such as Whisper \cite{radford2023robust}, Conformer \cite{gulati2020conformer}, and wav2vec 2.0 \cite{baevski2020wav2vec}—have demonstrated impressive performance on open-domain benchmarks, their reliability and generalizability in real-world procedural clinical environments remain insufficiently characterized \cite{mani2020asr}. Existing studies often rely on offline recordings, simulated noise, or single-center datasets, which fail to capture the performance degradation, institutional variability, and interaction dynamics encountered during routine endoscopic practice \cite{fan2022towards}.

A key bottleneck for advancing ASR-enabled Human–AI teaming in endoscopy is the limited availability of high-quality, domain-specific clinical speech data. Ethical considerations, patient privacy constraints, and the cost of expert annotation severely restrict large-scale data collection, particularly for intra-procedural speech \cite{nguyen2024exploring}. Consequently, there remains a critical evidence gap regarding whether ASR systems can maintain robust, real-time performance across different hospitals, endoscopy platforms, acoustic environments, and clinician speaking styles—an essential prerequisite for safe deployment of agent-based AI systems in clinical practice \cite{latif2020speech}.

To address these challenges, we present \textbf{EndoASR}, a fully open-source and domain-adapted Chinese ASR system specifically designed for intraprocedural dictation support in colonoscopy reporting workflows. To overcome the scarcity of domain-specific clinical speech data, we synthesize speech from structured endoscopy reports to construct a large-scale specialist training corpus, and further train a lightweight ASR model optimized for real-time, edge-deployed clinical environments. EndoASR is systematically evaluated using both retrospective single-center data collected from multiple experienced endoscopists and a prospective multi-center real-world clinical study spanning five medical centers and diverse clinical content categories across the endoscopic workflow. Beyond recognition accuracy, we further investigate the downstream workflow impact of EndoASR by integrating it with large language models for automatic clinical information extraction and AI-assisted communication. Together, these results position ASR as a critical entry point for reliable, real-world conversational AI systems in endoscopy units.

\begin{figure}[H]
\centering
\includegraphics[width=1.0\textwidth]{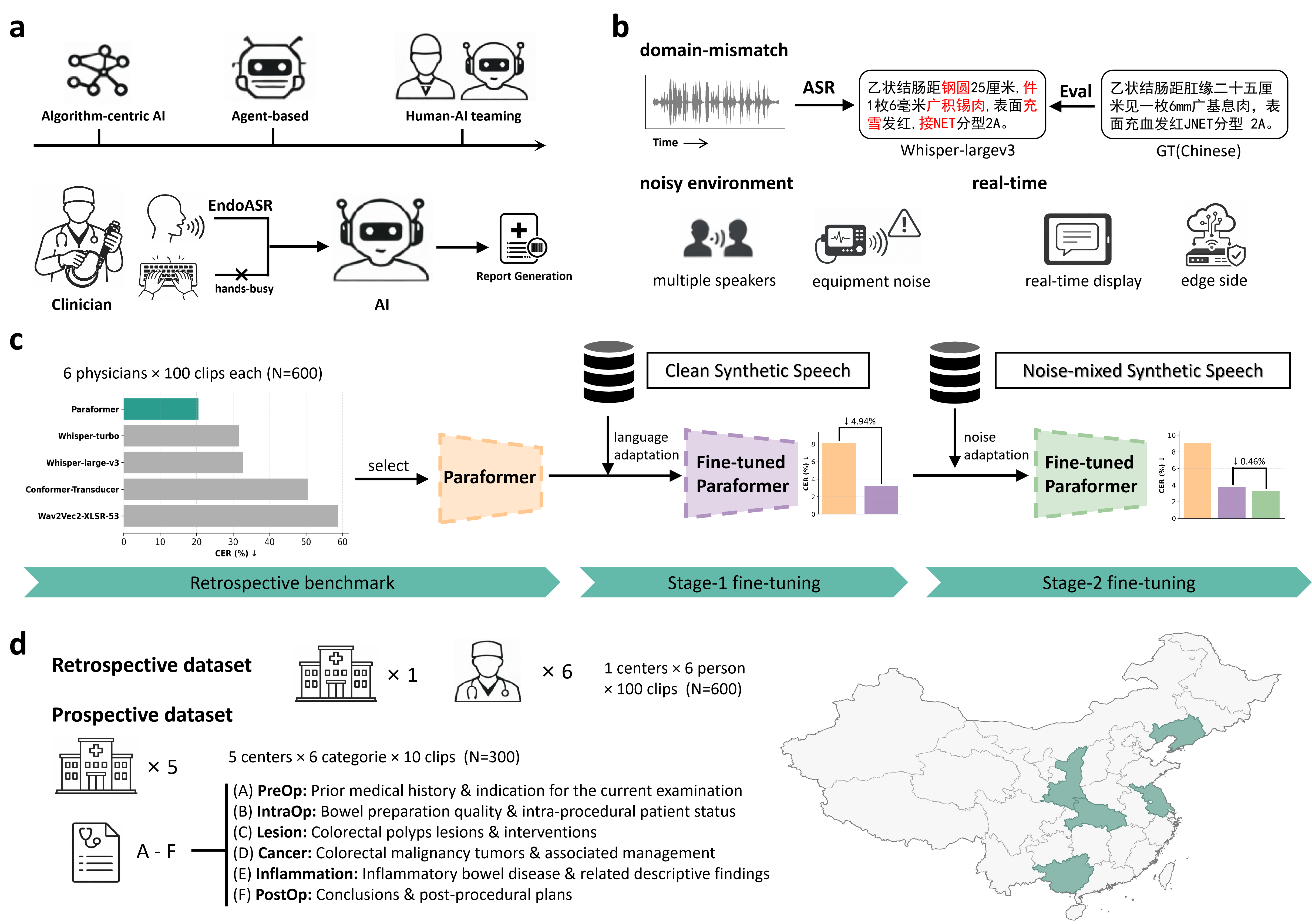}
\caption{
Overall framework of EndoASR for real-world deployment in gastrointestinal endoscopy.
The figure illustrates the conceptual and methodological design of EndoASR, spanning system motivation, model development, and clinical validation. Panel (a) presents the paradigm shift from algorithm-centric medical AI toward agent-based human–AI teaming, in which ASR serves as a real-time speech interface in hands-busy endoscopic workflows. Panel (b) summarizes the key challenges of speech recognition in endoscopy rooms, including specialized medical terminology, complex procedural acoustics, and real-time operational constraints. Panel (c) depicts the two-stage domain adaptation strategy, in which synthetic speech derived from structured endoscopy reports is used for domain-specific language adaptation, followed by noise-aware fine-tuning to improve robustness under realistic operating-room conditions. Panel (d) shows the progressive evaluation design, combining retrospective single-center validation across multiple endoscopists with prospective multi-center validation across diverse clinical content categories, enabling assessment of both controlled performance and real-world generalization.
}
\label{fig:framework}
\end{figure}

\section{Results}\label{sec2}
\subsection{Framework Overview}\label{subsec2_1}

To develop and validate a speech recognition system suitable for real-world endoscopic practice, we designed a unified framework that integrates domain adaptation, robustness enhancement, and progressive clinical validation. The overall framework is structured to support both controlled methodological development and rigorous assessment under realistic deployment conditions.

At the core of the framework, domain-specific language adaptation is addressed by leveraging structured endoscopy reports as a primary linguistic resource. By converting standardized clinical documentation into aligned speech–text data, the model is exposed to the terminology, phrasing patterns, and reporting conventions characteristic of gastrointestinal endoscopy. Building upon this linguistic adaptation, robustness to procedural acoustics is further enhanced through targeted training under noise conditions representative of endoscopy-room environments. This staged training strategy enables the model to disentangle adaptation to specialized medical language from adaptation to challenging acoustic conditions, while maintaining controllability during development.

Model evaluation is designed to progressively reflect real clinical usage. First, retrospective validation is conducted using intra-procedural speech collected from a single endoscopy center, involving multiple experienced endoscopists. This setting provides a controlled yet realistic benchmark to assess recognition accuracy across different speakers within a consistent institutional environment. Second, prospective evaluation is performed across multiple independent medical centers, with speech samples covering diverse clinical content categories spanning the endoscopic workflow. This multi-center, multi-category design enables assessment of cross-institutional generalization, acoustic variability, and linguistic diversity under real-world conditions.

Together, this framework supports systematic development, robustness analysis, and deployment-oriented evaluation of endoscopic ASR systems. An overview of the complete framework is illustrated in Fig.~\ref{fig:framework}.

\begin{table}[t]
\centering
\small
\renewcommand{\arraystretch}{1.08}

\caption{Performance comparison of general-purpose ASR models on the retrospective endoscopy dataset. Models were evaluated on 600 real intraoperative recordings from six endoscopists without domain-specific adaptation. Character error rate (CER), BLEU-1, and BERTScore are reported as mean $\pm$ standard deviation. Real-time factor (RTF) is included to illustrate model inference efficiency. Lower CER and RTF indicate better performance, while higher BLEU-1 and BERTScore reflect improved semantic consistency.}
\label{Tab_benchmark}
\begin{tabular*}{\textwidth}{@{\extracolsep{\fill}}lcccc@{}}
\toprule
\textbf{Model} &
\textbf{CER (\%)$\downarrow$} &
\textbf{BLEU-1 (\%)$\uparrow$} &
\textbf{BERTScore (\%)$\uparrow$} &
\textbf{RTF$\downarrow$} \\
\hline
Whisper-large-v3     & 32.75 $\pm$ 16.27 & 68.41 $\pm$ 15.38 & 84.21 $\pm$ 6.74 & 0.0531 \\
Whisper-turbo        & 31.63 $\pm$ 14.92 & 69.76 $\pm$ 13.27 & 83.72 $\pm$ 6.64 & 0.0182 \\
Conformer-Transducer & 50.51 $\pm$ 16.11 & 50.56 $\pm$ 16.03 & 72.45 $\pm$ 7.93 & 0.0230 \\
Wav2Vec2-XLSR-53     & 58.77 $\pm$ 14.23 & 44.83 $\pm$ 11.88 & 65.15 $\pm$ 6.99 & \textbf{0.0018} \\
Paraformer           & \textbf{20.52 $\pm$ 13.06} & \textbf{80.92 $\pm$ 10.26} & \textbf{90.11 $\pm$ 5.51} & 0.0051 \\
\hline
\end{tabular*}
\end{table}

\subsection{Retrospective Benchmarking of General-purpose ASR Models}\label{subsec2_2}
To establish a realistic baseline for endoscopic speech recognition, we benchmarked several widely used general-purpose ASR models on the retrospective clinical dataset, without any domain-specific adaptation. The evaluated models included Seaco-Paraformer~\cite{shi2023seaco}, Whisper-large-v3~\cite{radford2023robust}, Whisper-turbo, Conformer-Transducer (\url{https://huggingface.co/nvidia/stt_zh_conformer_transducer_large}), and wav2vec2-XLSR-53 (\url{https://huggingface.co/jonatasgrosman/wav2vec2-large-xlsr-53-chinese-zh-cn}), covering a range of contemporary ASR architectures with different trade-offs between accuracy, model size, and inference efficiency.

All models were evaluated on the same set of 600 real intraoperative recordings collected from six endoscopists. Performance was assessed using character error rate (CER), BLEU-1, and BERTScore, with manually verified reference transcripts. In addition, model parameter count and real-time factor (RTF) were reported to reflect computational complexity and deployment feasibility. No task-specific prompts, vocabulary constraints, or post-processing heuristics were applied, simulating a realistic out-of-the-box deployment scenario.

As summarized in Table~\ref{Tab_benchmark}, substantial performance differences were observed across models under endoscopy-room conditions. Among all evaluated systems, Paraformer achieved the best overall recognition performance, with the lowest CER (20.52\% $\pm$ 13.06\%) and the highest BLEU-1 (80.92\% $\pm$ 10.26\%) and BERTScore (90.11\% $\pm$ 5.51\%). In contrast, Whisper-large-v3 and Whisper-turbo exhibited higher CERs (32.75\% and 31.63\%, respectively), despite their strong performance on general-domain speech benchmarks. Conformer-Transducer and wav2vec2-XLSR-53 showed markedly degraded accuracy, with CERs exceeding 50\%, indicating limited robustness to intraoperative speech patterns and acoustic noise.

Notably, accuracy did not strictly correlate with model size or inference speed. While Whisper-large-v3 achieved competitive semantic similarity metrics, it required substantially more parameters (1.55B) and higher computational cost compared with Paraformer (220M). Conversely, wav2vec2-XLSR-53 achieved the lowest RTF (0.0018) but at the expense of poor transcription accuracy, highlighting the trade-off between efficiency and clinical usability.

Error analysis was conducted to qualitatively examine failure modes of general-purpose ASR models in endoscopic scenarios. Specifically, we selected four representative utterances that cover diverse anatomical locations, lesion morphologies, reporting conventions, and procedural expressions. The examples span multiple segments of the lower gastrointestinal tract (sigmoid colon, transverse colon, cecum, terminal ileum), include both protruded and flat lesions with standardized classifications (Paris, JNET), and incorporate structured procedural expressions such as BBPS scoring and withdrawal time. Fig.~\ref{fig:F2}b presents side-by-side comparisons of ground-truth transcriptions and predictions generated by Whisper-large-v3, the baseline Paraformer, and the proposed domain-adapted model (EndoASR-noise). For clarity, incorrectly recognized tokens are highlighted in red. Across these examples, general-purpose ASR models frequently exhibited errors in recognizing specialized medical terms. These qualitative observations complement the quantitative benchmark results and illustrate how domain mismatch in both vocabulary and reporting style contributes to recognition failures in general-purpose ASR systems.

Given that large multimodal or conversational foundation models (e.g., Kimi, Qwen3-Omni) incorporate ASR as a submodule within a broader language reasoning pipeline, their transcription outputs are influenced by downstream language modeling rather than pure speech recognition. As such, they were not included in the primary benchmark and are instead reported separately in the Supplementary Materials.

Overall, these results indicate that while general-purpose ASR models provide a useful baseline, their performance remains insufficient for reliable endoscopic workflow integration. The superior performance of Paraformer in this setting motivated its selection as the backbone for subsequent domain adaptation, which is described in the following section.

\begin{table}[t]
\centering
\small
\renewcommand{\arraystretch}{1.08}
\setlength{\tabcolsep}{4pt}

\caption{Performance of the two-stage fine-tuned ASR model on the synthetic endoscopic dataset. The baseline Paraformer model and the proposed model after stage-1 and stage-2 fine-tuning were evaluated on the synthetic test set, including both clean Mandarin speech and speech mixed with simulated endoscopy-room ambient noise. Results are reported as mean $\pm$ standard deviation. Lower CER and higher BLEU-1 and BERTScore indicate better performance.}
\label{Tab_synthetic}

\begin{tabular}{lcccc}
\toprule
\textbf{Condition} & \textbf{Model} & \textbf{CER (\%) $\downarrow$} & \textbf{BLEU-1 (\%) $\uparrow$} & \textbf{BERTScore (\%) $\uparrow$} \\
\midrule

\multirow{3}{*}{\textbf{Mandarin audio}}
& Paraformer     & 8.15 $\pm$ 4.43 & 91.99 $\pm$ 4.32 & 96.03 $\pm$ 2.03 \\
& EndoASR        & 3.71 $\pm$ 3.82 & 96.44 $\pm$ 3.60 & 98.60 $\pm$ 1.48 \\
& EndoASR-noise  & \textbf{3.21 $\pm$ 3.14} & \textbf{96.85 $\pm$ 3.06} & \textbf{98.81 $\pm$ 1.23} \\

\addlinespace[4pt]

\multirow{3}{*}{\textbf{Audio with noise}}
& Paraformer     & 9.10 $\pm$ 4.89 & 91.05 $\pm$ 4.79 & 95.49 $\pm$ 2.30 \\
& EndoASR        & 3.75 $\pm$ 3.77 & 96.39 $\pm$ 3.56 & 98.55 $\pm$ 1.47 \\
& EndoASR-noise  & \textbf{3.29 $\pm$ 3.28} & \textbf{96.79 $\pm$ 3.16} & \textbf{98.78 $\pm$ 1.25} \\

\bottomrule
\end{tabular}
\end{table}

\subsection{Domain-adapted EndoASR: Two-stage Fine-tuning and Performance}\label{subsec2_3}
Building on the retrospective benchmark results, we selected Paraformer as the backbone architecture for domain adaptation and developed a two-stage fine-tuning strategy to address the linguistic and acoustic challenges of endoscopic speech. An overview of the fine-tuning pipeline is illustrated in Fig.~\ref{fig:framework}c.

Importantly, the two-stage adaptation process exclusively leveraged the synthetic datasets: Stage-1 fine-tuning was performed using clean synthetic speech to adapt domain-specific language patterns, while Stage-2 fine-tuning utilized noise-augmented synthetic speech to enhance acoustic robustness. All retrospective and prospective real-world datasets were strictly reserved for evaluation and were not used during model training or adaptation.

We first evaluated the effectiveness of domain adaptation on the synthetic endoscopic dataset. As shown in Table~\ref{Tab_synthetic}, fine-tuning the baseline Paraformer model on report-level synthetic speech led to a substantial performance improvement. Compared with the original model, the EndoASR model reduced the CER from 8.15\% to 3.71\% on clean synthetic speech and from 9.10\% to 3.75\% under noise-mixed conditions, accompanied by consistent gains in BLEU-1 and BERTScore. These improvements indicate enhanced recognition of domain-specific vocabulary and reporting-style expressions.

The second-stage fine-tuning further improved robustness to realistic acoustic conditions. After incorporating noise-aware adaptation, the EndoASR-noise model achieved the best overall performance on the synthetic test set, with CER decreasing to 3.21\% on clean speech and 3.29\% on noise-mixed speech. The marginal but consistent gains across all metrics suggest that the two-stage strategy effectively captured both linguistic specialization and noise characteristics relevant to endoscopy-room environments.

\begin{figure}[H]
\centering
\includegraphics[width=1.0\textwidth]{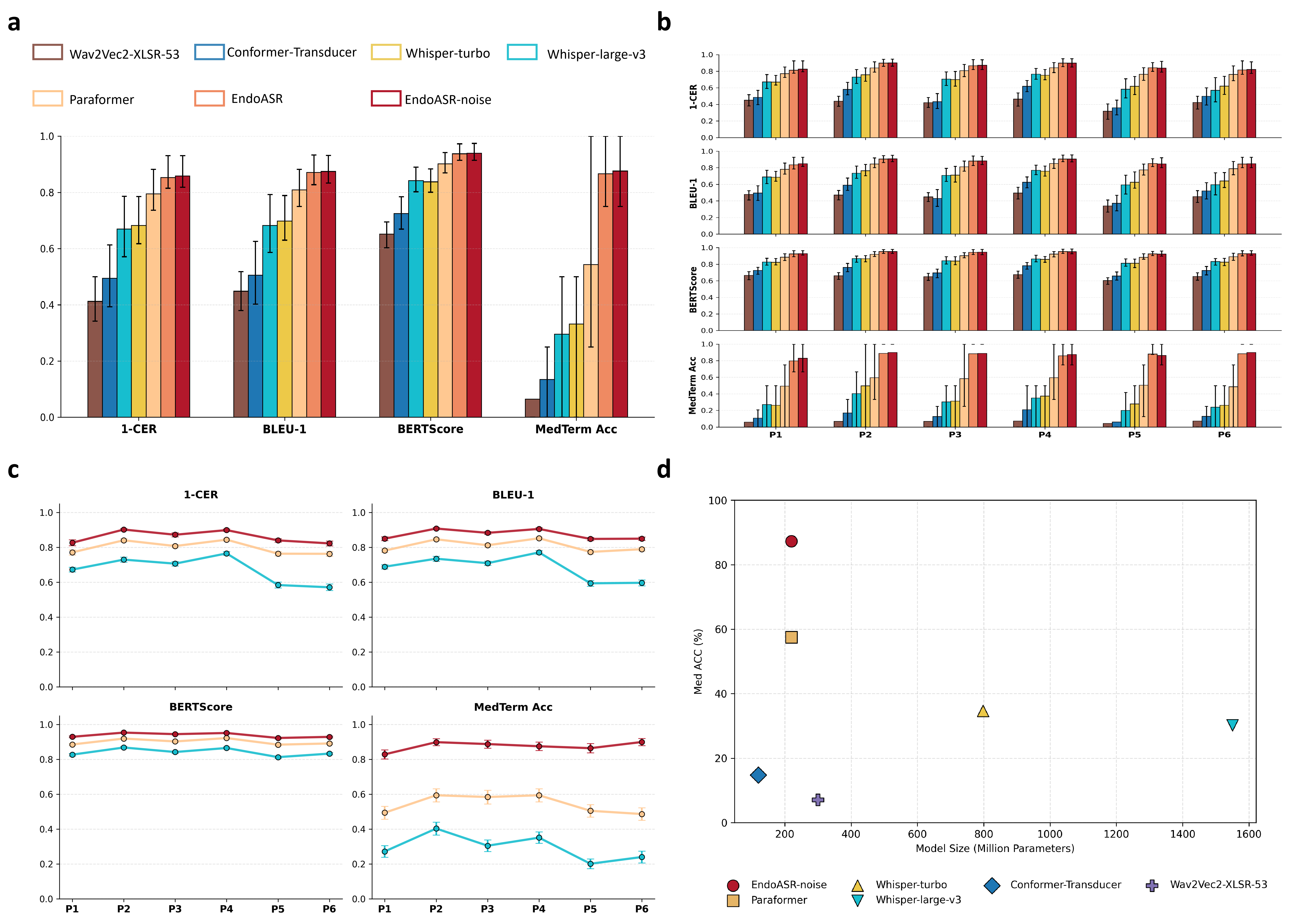}
\caption{
Retrospective evaluation of ASR performance, inter-speaker variability, and noise-related robustness analysis.
Panel (a) compares the performance of different ASR models on the retrospective clinical dataset across four metrics, including 1–CER, BLEU-1, BERTScore, and medical terminology accuracy. Panel (b) reports model performance stratified by individual endoscopists (P1–P6) on the same retrospective dataset, highlighting speaker-dependent variations across evaluation metrics. Panel (c) further visualizes inter-speaker variability by summarizing performance differences across endoscopists, illustrating the heterogeneity of intra-procedural speech characteristics. Panel (d) illustrates the relationship between model parameter size and medical terminology accuracy across different methods. Models closer to the upper-left corner achieve a more favorable efficiency–accuracy trade-off, delivering higher terminology accuracy with fewer parameters.
}
\label{fig:retrospective}
\end{figure}

We next evaluated the domain-adapted models on the retrospective real-world dataset comprising 600 intraoperative recordings from six endoscopists. As summarized in Table~\ref{Tab_benchmark_ours}, both stages of fine-tuning consistently outperformed the baseline Paraformer model across all physicians. On average, the EndoASR-noise model reduced CER from 20.52\% to 14.14\%, representing a relative error reduction of approximately 31\%, while simultaneously improving BLEU-1 and BERTScore.

Performance gains were observed across all six endoscopists, despite inter-physician variability in speaking style, speech rate, and acoustic conditions. In most cases, stage-2 fine-tuning yielded the lowest CER and highest semantic similarity scores, indicating improved generalization to real intraoperative speech. Notably, for some physicians (e.g., P4 and P5), EndoASR and EndoASR-noise models exhibited comparable performance, suggesting that language adaptation contributed the majority of gains, while noise adaptation provided additional but more modest improvements.

As shown in Fig.~\ref{fig:retrospective}, the performance of different ASR models on the retrospective clinical dataset highlights the impact of domain adaptation and robustness enhancement. Overall, these results demonstrate that domain-specific fine-tuning substantially enhances ASR performance in endoscopic settings and that a two-stage strategy effectively balances linguistic specialization and acoustic robustness. These findings motivated further validation of the proposed model in prospective multi-center clinical deployments, as described in the following section.

\begin{table}[t]
\centering
\footnotesize
\renewcommand{\arraystretch}{1.02}
\setlength{\tabcolsep}{2.5pt}

\caption{Retrospective evaluation of the baseline and domain-adapted ASR models across six endoscopists. Performance of the baseline Paraformer model and the proposed models after stage-1 and stage-2 fine-tuning was evaluated on 600 real intraoperative recordings from six endoscopists (P1--P6). Results are reported as mean $\pm$ standard deviation for CER, BLEU-1, BERTScore, and medical term accuracy (Med ACC), all in \%. Lower CER and higher BLEU-1, BERTScore, and Med ACC indicate better performance.}
\label{Tab_benchmark_ours}

\begin{tabular*}{\textwidth}{@{\extracolsep{\fill}}l l c c c c @{}}
\toprule
& \textbf{Model} & \textbf{CER $\downarrow$} & \textbf{BLEU-1 $\uparrow$} & \textbf{BERTScore $\uparrow$} & \textbf{Med ACC $\uparrow$} \\
\midrule

\multirow{3}{*}{P1}
& Paraformer     & 25.06 $\pm$ 18.86 & 78.14 $\pm$ 10.94 & 88.56 $\pm$ 5.51 & 49.41 $\pm$ 36.33 \\
& EndoASR        & 20.88 $\pm$ 22.13 & 83.61 $\pm$ 12.38 & 92.43 $\pm$ 5.43 & 79.71 $\pm$ 29.78 \\
& EndoASR-noise  & \textbf{18.51 $\pm$ 19.57} & \textbf{84.99 $\pm$ 11.09} & \textbf{92.97 $\pm$ 5.03} & \textbf{82.91 $\pm$ 26.14} \\
\addlinespace[1pt]

\multirow{3}{*}{P2}
& Paraformer     & 15.93 $\pm$ 9.44 & 84.62 $\pm$ 8.85 & 91.96 $\pm$ 4.87 & 59.42 $\pm$ 37.22 \\
& EndoASR        & 10.02 $\pm$ 6.48 & 90.55 $\pm$ 5.95 & 95.28 $\pm$ 3.39 & 88.88 $\pm$ 21.67 \\
& EndoASR-noise  & \textbf{9.77 $\pm$ 6.25} & \textbf{90.79 $\pm$ 5.71} & \textbf{95.38 $\pm$ 3.29} & \textbf{89.89 $\pm$ 19.75} \\
\addlinespace[1pt]

\multirow{3}{*}{P3}
& Paraformer     & 19.24 $\pm$ 9.75 & 81.21 $\pm$ 9.44 & 90.30 $\pm$ 5.39 & 58.41 $\pm$ 39.00 \\
& EndoASR        & 13.21 $\pm$ 11.83 & 88.05 $\pm$ 8.43 & 94.42 $\pm$ 4.25 & 88.46 $\pm$ 24.29 \\
& EndoASR-noise  & \textbf{12.76 $\pm$ 11.26} & \textbf{88.32 $\pm$ 8.17} & \textbf{94.49 $\pm$ 4.18} & \textbf{88.80 $\pm$ 23.06} \\
\addlinespace[1pt]

\multirow{3}{*}{P4}
& Paraformer     & 15.59 $\pm$ 7.75 & 85.21 $\pm$ 7.42 & 92.24 $\pm$ 4.00 & 59.42 $\pm$ 37.24 \\
& EndoASR        & 10.13 $\pm$ 7.04 & 90.52 $\pm$ 6.45 & \textbf{95.29 $\pm$ 3.59} & 86.02 $\pm$ 25.26 \\
& EndoASR-noise  & \textbf{10.09 $\pm$ 7.07} & \textbf{90.57 $\pm$ 6.59} & 95.15 $\pm$ 3.79 & \textbf{87.54 $\pm$ 24.54} \\
\addlinespace[1pt]

\multirow{3}{*}{P5}
& Paraformer     & 23.63 $\pm$ 12.02 & 77.38 $\pm$ 11.13 & 88.45 $\pm$ 6.20 & 50.50 $\pm$ 35.72 \\
& EndoASR        & \textbf{15.66 $\pm$ 9.74} & \textbf{85.19 $\pm$ 9.02} & \textbf{92.46 $\pm$ 5.04} & \textbf{88.13 $\pm$ 24.53} \\
& EndoASR-noise  & 16.00 $\pm$ 9.89 & 84.83 $\pm$ 9.17 & 92.29 $\pm$ 5.02 & 86.44 $\pm$ 26.68 \\
\addlinespace[1pt]

\multirow{3}{*}{P6}
& Paraformer     & 23.70 $\pm$ 13.86 & 78.94 $\pm$ 10.47 & 89.14 $\pm$ 5.54 & 48.65 $\pm$ 35.68 \\
& EndoASR        & 18.40 $\pm$ 15.29 & 84.61 $\pm$ 10.58 & 92.68 $\pm$ 4.95 & 88.46 $\pm$ 24.06 \\
& EndoASR-noise  & \textbf{17.72 $\pm$ 13.72} & \textbf{84.98 $\pm$ 9.84} & \textbf{92.92 $\pm$ 4.85} & \textbf{89.98 $\pm$ 20.85} \\
\addlinespace[1pt]

\multirow{3}{*}{Mean}
& Paraformer     & 20.52 $\pm$ 13.06 & 80.92 $\pm$ 10.26 & 90.11 $\pm$ 5.51 & 54.30 $\pm$ 37.20 \\
& EndoASR        & 14.72 $\pm$ 13.83 & 87.09 $\pm$ 9.50 & 93.76 $\pm$ 4.68 & 86.61 $\pm$ 25.26 \\
& EndoASR-noise  & \textbf{14.14 $\pm$ 12.63} & \textbf{87.41 $\pm$ 9.01} & \textbf{93.87 $\pm$ 4.57} & \textbf{87.59 $\pm$ 23.77} \\

\bottomrule
\end{tabular*}
\end{table}

\begin{figure}[t]
\centering
\includegraphics[width=1.0\textwidth]{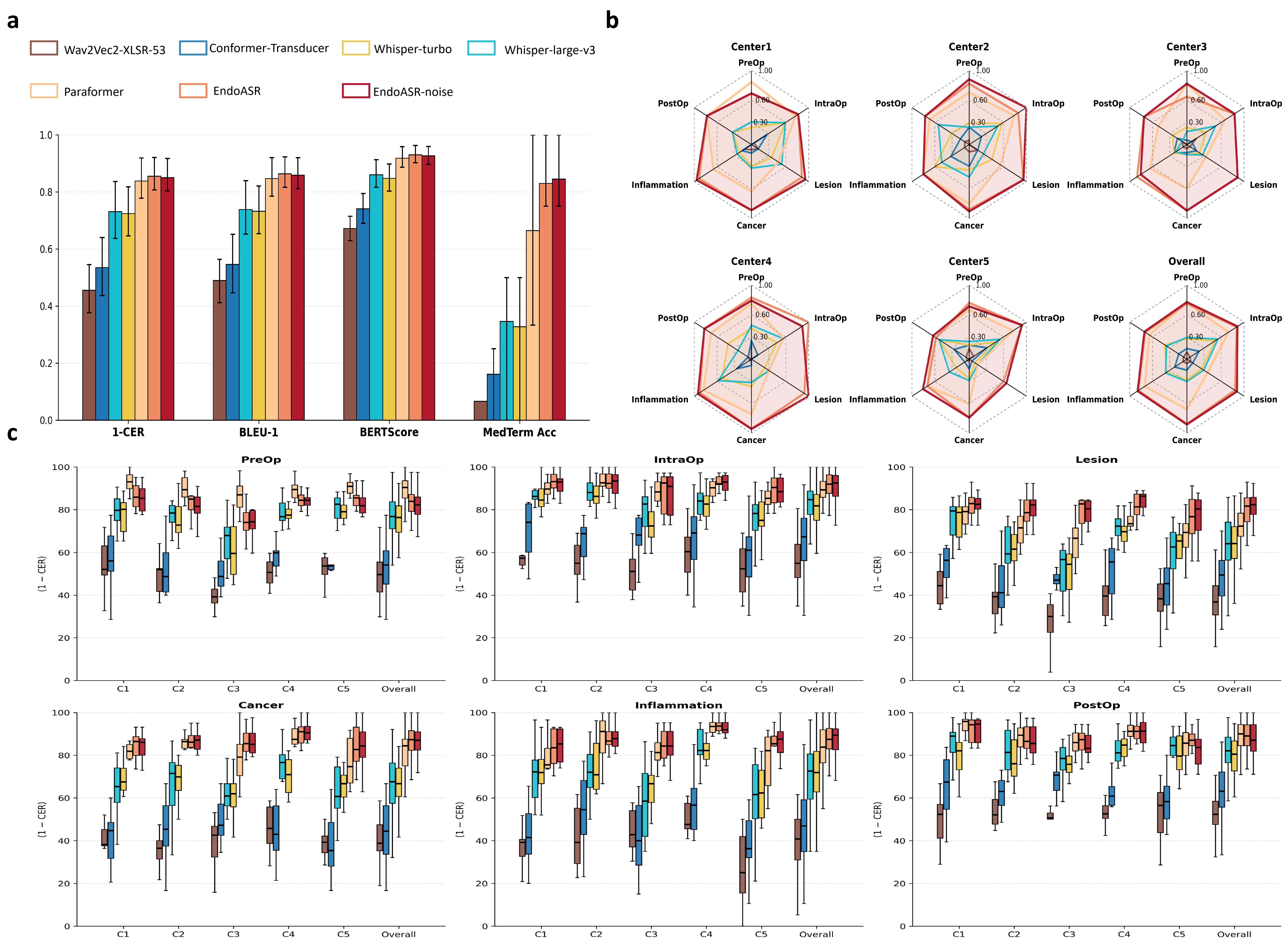}
\caption{
Prospective multi-center evaluation across clinical content categories.
Panel (a) summarizes ASR performance on the prospective multi-center dataset across evaluation metrics, reflecting overall recognition quality under real-world clinical conditions. Panel (b) presents medical terminology accuracy stratified by center and content category, highlighting variation in domain-specific term recognition. Panel (c) compares 1–CER performance of different ASR models across centers and clinical categories, illustrating cross-institutional and content-dependent performance patterns in prospective evaluation.
}
\label{fig:prospective}
\end{figure}

\subsection{Prospective Multi-center Real-world Evaluation}\label{subsec2_4}

To further assess the real-world clinical utility of the proposed endoscopic ASR system, we conducted a prospective multi-center evaluation across five independent endoscopy centers. This dataset consisted of 300 intraoperative recordings collected during routine clinical practice, with each center contributing 60 unique audio--text pairs. Compared with retrospective data, the prospective recordings better reflect real deployment conditions, including spontaneous speech, workflow interruptions, and institutional variability.

Beyond reporting the number of participating centers, we further characterized the prospective dataset from an acoustic perspective. Specifically, we extracted Mel-frequency cepstral coefficients (MFCCs) from each utterance to capture short-term spectral envelope characteristics of the speech signal, which are widely used to represent acoustic structure while suppressing fine-grained waveform variability.

\begin{figure}[t]
\centering
\includegraphics[width=1.0\textwidth]{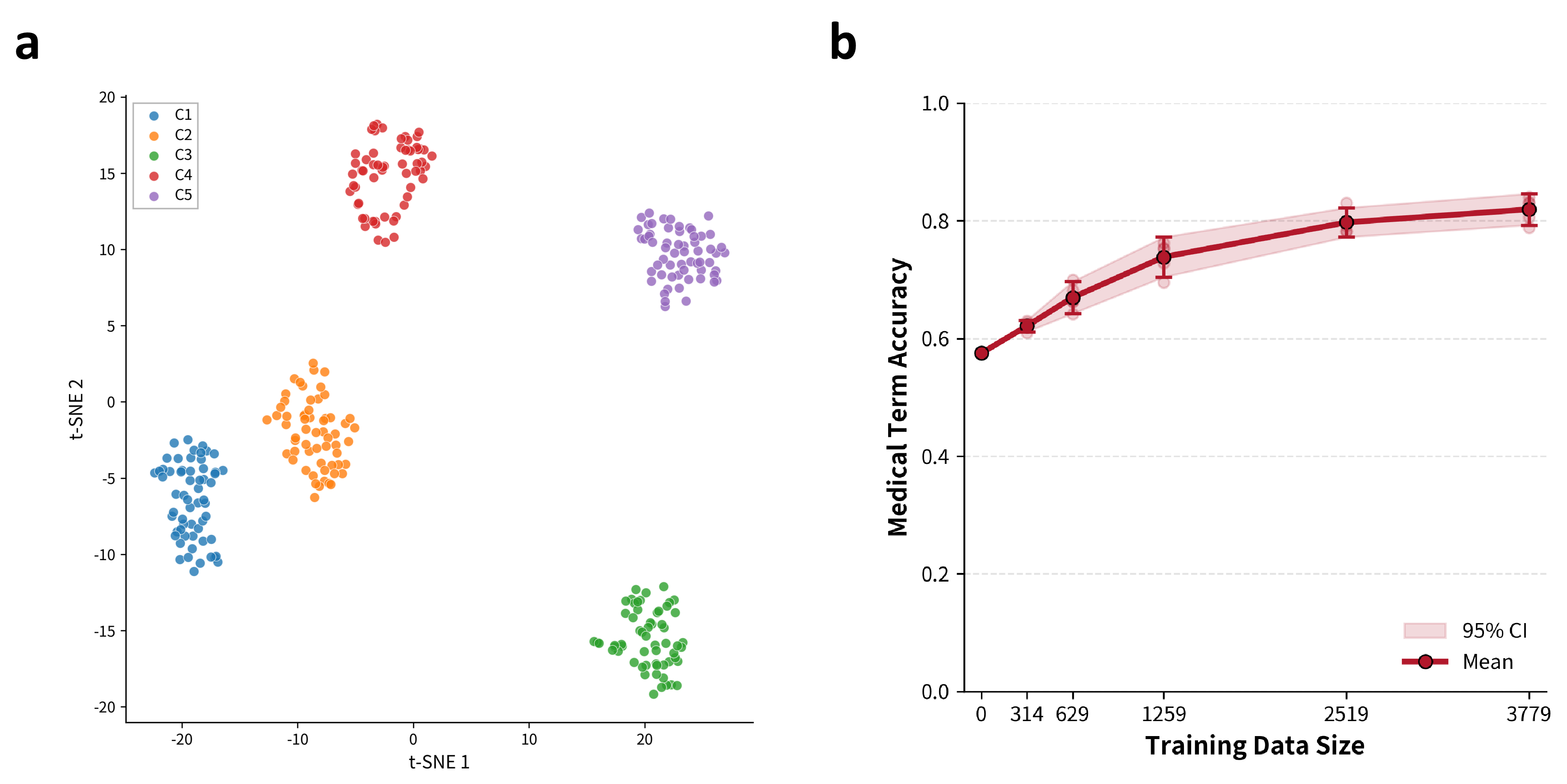}
\caption{
Acoustic variability across centers and the impact of synthetic data scaling on medical term recognition.
(a) t-SNE visualization of MFCC-based acoustic embeddings from the prospective multi-center dataset. Each point represents a speech segment, with colors indicating different clinical centers. Distinct clustering patterns are observed across centers, reflecting substantial inter-center variability in acoustic characteristics, likely due to differences in recording environments, speakers, and procedural conditions.
(b) Effect of synthetic training data scale on medical term recognition accuracy. Performance is plotted against the size of synthetic data used for domain adaptation, where the point at size 0 corresponds to the baseline model without adaptation (Paraformer). Results show a consistent improvement in accuracy as the amount of synthetic data increases.
}
\label{fig:F5}
\end{figure}

Fig.~\ref{fig:F5}a visualizes the MFCC representations using t-SNE. Each point corresponds to an utterance and is colored by clinical center. As shown in Fig.~\ref{fig:F5}a, recordings from different centers form clearly separated clusters in the acoustic embedding space.

This pattern indicates substantial inter-center variability in speech characteristics. Such differences may arise from a combination of factors, including recording environments, background noise conditions, speaker behavior, and clinical workflow differences. Although the dataset is collected under a unified task setting, the acoustic properties of the recordings remain center-dependent.

These findings highlight the importance of evaluating ASR systems under multi-center conditions that reflect realistic acoustic heterogeneity, rather than assuming a single-center or acoustically homogeneous deployment scenario.

As shown in Table~\ref{Tab_prospective}, both domain-adapted models consistently outperformed the baseline Paraformer across all centers on conventional ASR metrics, including CER, BLEU-1, and BERTScore. Overall, the two-stage adapted models demonstrated stable and reproducible gains under prospective real-world conditions, indicating that the improvements observed in retrospective evaluations largely translated to routine clinical deployment.

However, conventional character-level metrics alone do not fully capture clinical usefulness in endoscopic practice. To better reflect domain-specific performance, we further evaluated medical terminology accuracy (Med ACC) on the prospective dataset. A curated set of clinically relevant terms was manually identified and verified by participating endoscopists, serving as a clinically grounded reference for evaluation.

The EndoASR-noise model consistently outperformed EndoASR in four out of five centers and achieved superior overall performance across the full cohort. Specifically, the overall medical terminology accuracy improved from 82.27\% with EndoASR to 84.16\% with EndoASR-noise, indicating that noise-aware adaptation provides robust gains under diverse clinical acoustic conditions.

These results indicate that while overall CER differences between EndoASR and EndoASR-noise models were modest in the prospective setting, noise-aware fine-tuning in EndoASR-noise provided tangible benefits for the accurate recognition of clinically critical terminology. Given that correct transcription of medical terms directly affects downstream documentation, decision support, and patient safety, this improvement is particularly meaningful for real-world clinical deployment.

Taken together, the prospective multi-center evaluation demonstrates that the proposed two-stage adaptation strategy not only generalizes across institutions, but also enhances recognition of domain-specific medical terminology that is essential for endoscopic reporting and clinical decision-making.

Fig.~\ref{fig:prospective}b and Fig.~\ref{fig:prospective}c jointly illustrate model performance across six clinical content categories, covering the full endoscopic workflow from pre-procedural assessment to post-procedural conclusions. While character error rate (CER) remains an important global indicator of transcription quality, category-wise analysis reveals that CER differences between the two domain-adapted models are relatively modest in the prospective setting.

Across intra-procedural and lesion-focused categories, including bowel preparation assessment, colorectal polyp characterization, malignancy-related descriptions, and inflammatory bowel disease findings (IntraOp, Lesion, Cancer and Inflammation), both domain-adapted models consistently outperform general-purpose ASR systems in terms of CER. However, within these categories, the performance gap between EndoASR and EndoASR-noise models is small, reflecting the heterogeneous acoustic conditions present in prospective multi-center data.

In contrast, analysis of medical terminology accuracy (Fig.~\ref{fig:prospective}b) reveals a clearer and more clinically meaningful distinction between the two adaptation stages. In these same intra-procedural and lesion-related categories, the EndoASR-noise model achieves consistently higher terminology accuracy across centers, indicating that noise-aware adaptation improves the recognition stability of clinically critical terms even when overall CER differences are subtle.

Taken together, these results highlight that medical terminology accuracy constitutes the most clinically meaningful indicator of ASR performance in endoscopic workflows. As demonstrated in Fig.~\ref{fig:prospective}b, the proposed domain-adapted models substantially outperform general-purpose ASR systems across all six content categories, with particularly pronounced gains in lesion-related and intra-procedural scenarios where precise recognition of specialized terminology is critical.

The consistent advantage of the two-stage adaptation strategy over baseline and general ASR models indicates effective domain specialization, enabling robust recognition of endoscopy-specific terms that are frequently misrecognized by out-of-domain systems. Compared with conventional character-level metrics, terminology accuracy more directly reflects clinical usability, as errors in key medical terms can fundamentally alter clinical interpretation even when overall CER differences are small.

These findings demonstrate that the primary value of the proposed domain-adapted ASR system lies in its ability to reliably capture domain-specific medical vocabulary, thereby providing a stronger foundation for accurate clinical documentation and downstream AI-assisted analysis in real-world endoscopic practice.

\begin{table}[t]
\centering
\footnotesize
\renewcommand{\arraystretch}{1.02}
\setlength{\tabcolsep}{2.5pt}

\caption{Prospective multi-center evaluation of the domain-adapted ASR models in real-world endoscopic practice. Performance of the baseline Paraformer model and the proposed domain-adapted models after stage-1 and stage-2 fine-tuning was evaluated on a prospective multi-center dataset collected from five independent endoscopy centers (C1--C5). Each center contributed 60 intraoperative recordings spanning six clinical content categories. Results are reported as mean $\pm$ standard deviation for CER, BLEU-1, BERTScore, and medical term accuracy (Med ACC), all in \%. Lower CER and higher BLEU-1, BERTScore, and Med ACC indicate better performance.}
\label{Tab_prospective}

\begin{tabular*}{\textwidth}{@{\extracolsep{\fill}}l l c c c c @{}}
\toprule
& \textbf{Model} & \textbf{CER $\downarrow$} & \textbf{BLEU-1 $\uparrow$} & \textbf{BERTScore $\uparrow$} & \textbf{Med ACC $\uparrow$} \\
\midrule

\multirow{3}{*}{C1}
& Paraformer     & 15.24 $\pm$ 12.28 & 85.91 $\pm$ 9.36 & 92.56 $\pm$ 4.80 & 66.72 $\pm$ 36.53 \\
& EndoASR        & \textbf{14.50 $\pm$ 11.42} & \textbf{86.89 $\pm$ 8.67} & \textbf{93.26 $\pm$ 4.10} & 86.22 $\pm$ 25.38 \\
& EndoASR-noise  & 14.75 $\pm$ 10.81 & 86.49 $\pm$ 8.34 & 92.90 $\pm$ 4.05 & \textbf{89.05 $\pm$ 21.87} \\
\addlinespace[1pt]

\multirow{3}{*}{C2}
& Paraformer     & 14.76 $\pm$ 11.16 & 85.78 $\pm$ 10.59 & 92.84 $\pm$ 5.11 & 67.81 $\pm$ 38.41 \\
& EndoASR        & \textbf{14.30 $\pm$ 8.62} & \textbf{86.37 $\pm$ 7.96} & \textbf{93.16 $\pm$ 4.35} & 87.16 $\pm$ 22.44 \\
& EndoASR-noise  & 14.83 $\pm$ 9.28 & 85.78 $\pm$ 8.74 & 92.64 $\pm$ 4.73 & \textbf{89.72 $\pm$ 21.10} \\
\addlinespace[1pt]

\multirow{3}{*}{C3}
& Paraformer     & 19.26 $\pm$ 11.47 & 82.50 $\pm$ 10.67 & 90.68 $\pm$ 5.50 & 56.10 $\pm$ 37.84 \\
& EndoASR        & \textbf{16.99 $\pm$ 9.24} & \textbf{84.66 $\pm$ 8.34} & \textbf{91.97 $\pm$ 4.47} & \textbf{84.37 $\pm$ 27.76} \\
& EndoASR-noise  & 17.42 $\pm$ 9.37 & 84.14 $\pm$ 8.49 & 91.81 $\pm$ 4.48 & 84.21 $\pm$ 29.30 \\
\addlinespace[1pt]

\multirow{3}{*}{C4}
& Paraformer     & 12.34 $\pm$ 8.10 & 87.97 $\pm$ 7.61 & 93.69 $\pm$ 4.50 & 66.44 $\pm$ 40.76 \\
& EndoASR        & 11.20 $\pm$ 6.55 & 89.01 $\pm$ 6.39 & \textbf{94.44 $\pm$ 3.51} & 85.83 $\pm$ 30.34 \\
& EndoASR-noise  & \textbf{10.97 $\pm$ 6.07} & \textbf{89.26 $\pm$ 5.96} & 94.31 $\pm$ 3.73 & \textbf{88.07 $\pm$ 27.57} \\
\addlinespace[1pt]

\multirow{3}{*}{C5}
& Paraformer     & 19.38 $\pm$ 12.00 & 81.01 $\pm$ 11.70 & 89.38 $\pm$ 6.69 & 57.33 $\pm$ 35.82 \\
& EndoASR        & \textbf{15.60 $\pm$ 9.38} & \textbf{84.97 $\pm$ 8.64} & \textbf{91.90 $\pm$ 4.54} & 68.48 $\pm$ 36.98 \\
& EndoASR-noise  & 16.89 $\pm$ 9.45 & 83.64 $\pm$ 8.87 & 91.36 $\pm$ 4.61 & \textbf{70.45 $\pm$ 35.16} \\
\addlinespace[1pt]

\multirow{3}{*}{Mean}
& Paraformer     & 16.20 $\pm$ 11.43 & 84.63 $\pm$ 10.39 & 91.83 $\pm$ 5.60 & 61.63 $\pm$ 38.18 \\
& EndoASR        & \textbf{14.52 $\pm$ 9.37} & \textbf{86.38 $\pm$ 8.20} & \textbf{92.95 $\pm$ 4.31} & 82.27 $\pm$ 29.95 \\
& EndoASR-noise  & 14.97 $\pm$ 9.41 & 85.86 $\pm$ 8.39 & 92.60 $\pm$ 4.45 & \textbf{84.16 $\pm$ 28.52} \\

\bottomrule
\end{tabular*}
\end{table}

\subsection{Impact on Real-world Clinical Workflow and Clinician--AI Interaction}
\label{subsec2_5}

Beyond standalone speech recognition accuracy, we further examined the impact of domain-adapted ASR on real-world clinical workflows through an end-to-end clinician--AI interaction scenario. Rather than treating ASR as an isolated component, this experiment situates speech recognition within a complete interaction loop, reflecting emerging agent-based paradigms for clinical documentation and decision support in procedural medicine.

Specifically, we constructed a live demonstration in which spoken intraoperative utterances served as the sole input modality, and downstream outputs were generated by a large language model (LLM). Two ASR configurations were compared side by side under identical downstream conditions: (1) the default ASR module provided by the LLM system (based on Whisper), and (2) the proposed domain-adapted EndoASR module. The LLM architecture, prompting strategy, and information extraction pipeline were held constant, thereby isolating the effect of ASR quality on clinician--AI interaction.

\begin{figure}[H]
\centering
\includegraphics[width=1.0\textwidth]{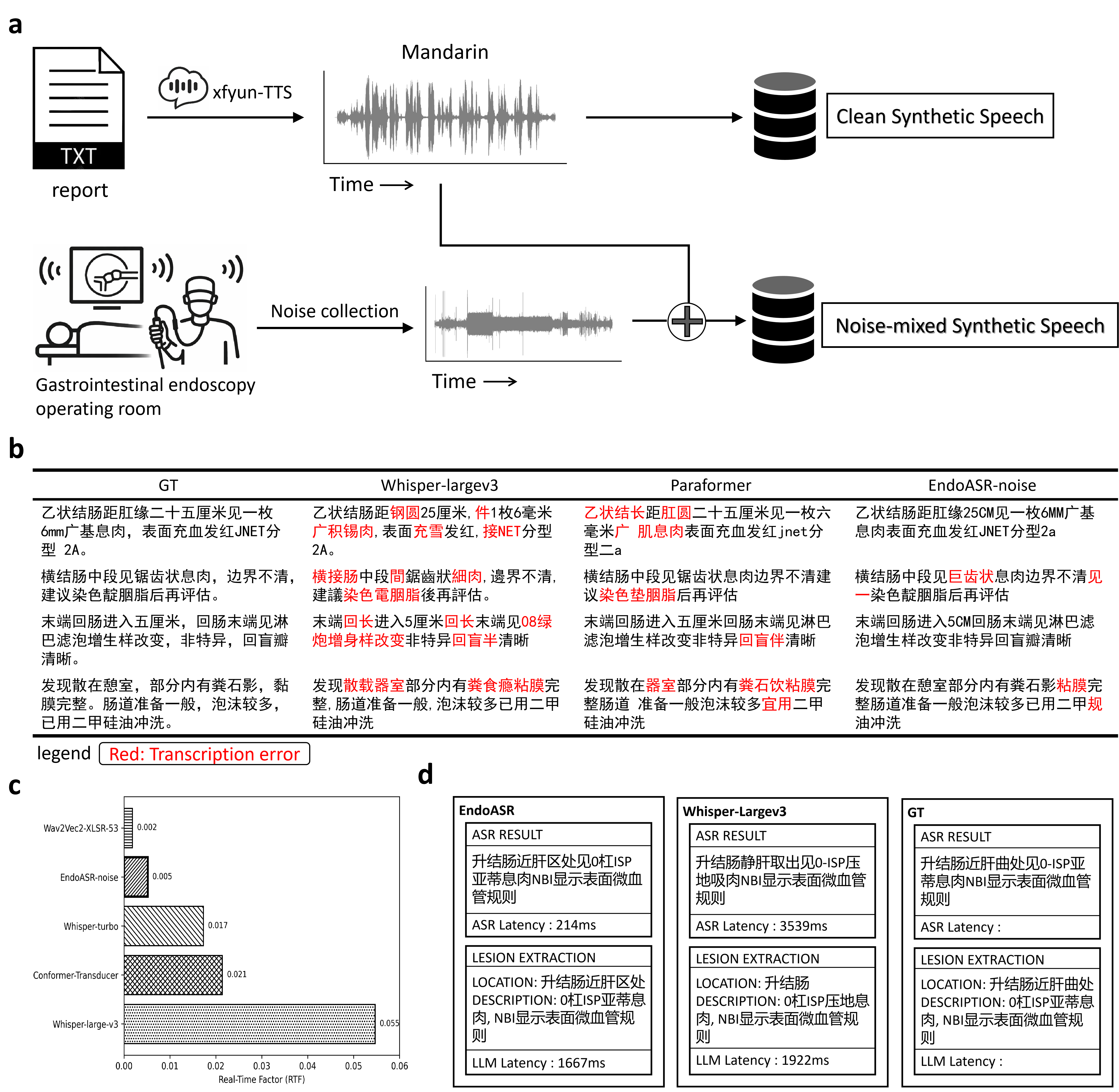}
\caption{
Data construction, transcription quality, runtime efficiency, and downstream usability of EndoASR.
Panel (a) illustrates the construction of the training data, including synthetic speech generated from structured clinical text and noise-augmented speech informed by real endoscopy-room acoustics. Panel (b) compares representative transcription outputs produced by Whisper-large-v3, Paraformer, and EndoASR-noise, highlighting differences in domain-specific terminology recognition and transcription errors (shown in red). Panel (c) presents a comparison of real-time factor (RTF) across different ASR models, reflecting their runtime efficiency under practical deployment settings. Panel (d) demonstrates downstream usage by integrating ASR outputs with a large language model for structured clinical information extraction, comparing the performance of EndoASR and Whisper-large-v3 in supporting accurate and usable endoscopy report generation.
}
\label{fig:F2}
\end{figure}

In both settings, clinicians dictated free-form intraoperative descriptions, and the LLM was tasked with extracting structured clinical information, including lesion location, descriptive findings, and procedural actions(Fig.~\ref{fig:F2}d). A key distinction lies in system responsiveness: the proposed EndoASR produces transcription outputs almost immediately after speech input, with visible text appearing on the screen within approximately 200~ms after utterance completion. This near real-time feedback enables fluid interaction , which is critical in high-acuity endoscopic workflows.

The demonstration focuses on clinically meaningful behaviors rather than conventional ASR metrics alone. We qualitatively examine transcription fidelity of domain-specific terminology, handling of fragmented and spontaneous speech, and the robustness of downstream structured information extraction. Particular attention is paid to error propagation patterns, where minor ASR inaccuracies lead to incorrect or ambiguous structured outputs, despite otherwise strong language model capabilities.

Qualitative comparisons reveal that replacing the default ASR with the proposed EndoASR results in more accurate transcription of medical terminology, fewer downstream extraction errors, and more coherent structured reports generated by the LLM. These improvements are especially pronounced for lesion descriptions, anatomical locations, and procedural actions, which are poorly represented in general-purpose ASR vocabularies but are critical for endoscopic documentation.

To illustrate these effects in a realistic and interpretable manner, we provide a side-by-side video demonstration in the Supplementary Material. The left panel shows clinician--AI interaction using the proposed EndoASR module, while the right panel uses the default ASR module provided by the LLM system, enabling direct visual comparison of transcription latency, textual accuracy, and downstream structured outputs in real time. By embedding ASR within a complete clinical interaction loop, this experiment demonstrates that domain-adapted speech recognition is not merely a technical refinement, but a critical enabling component for safe, efficient, and trustworthy AI-assisted workflows in gastrointestinal endoscopy.

\section{Discussion}\label{sec3}

This study addresses the out-of-distribution (OOD) challenge of applying general-purpose automatic speech recognition (ASR) systems to gastrointestinal endoscopy, where speech is characterized by domain-specific terminology and complex acoustic conditions. We first conducted a systematic benchmark on retrospective data to identify a strong foundation model, and subsequently developed a two-stage adaptation strategy using synthetic data, targeting domain-specific language modeling and noise robustness, respectively.

Crucially, the proposed model was evaluated on a prospective multi-center dataset, demonstrating consistent performance across heterogeneous clinical environments. This multi-center validation highlights both the effectiveness of the adaptation strategy and the practical reliability of the system under real-world deployment conditions.

In contrast to existing medical ASR efforts such as MedASR, which primarily focus on English, our work provides a dedicated solution for Chinese clinical speech, addressing a critical gap in current medical AI infrastructure.

\textbf{Synthetic data scaling exhibits diminishing returns beyond moderate data sizes.}
We further investigated the impact of synthetic training data scale on domain adaptation performance. As shown in Fig.~\ref{fig:F5}b, increasing the amount of synthetic data consistently improves medical term recognition accuracy compared to the baseline model without adaptation. However, the performance gains gradually saturate as the data size increases. Notably, substantial improvements are achieved with the initial increase in data scale, whereas further expansion yields only marginal benefits.

This trend suggests that synthetic data primarily serves to bridge the domain gap between general-purpose ASR and endoscopic speech, and that this gap can be largely mitigated with a moderate amount of domain-relevant data. Beyond this point, the remaining errors are more likely attributed to factors such as pronunciation variability and acoustic diversity, which cannot be fully addressed by simply increasing data volume~\cite{winata2020learning, qian2022layer, huang2004accent, bell2020adaptation}.

\textbf{Stage-1 versus stage-2 performance in prospective multi-center evaluation.}
In the prospective multi-center setting, stage-1 achieved slightly lower CER than stage-2 across most centers and content categories (Fig.~\ref{fig:prospective}c). This observation does not negate the value of noise-aware adaptation, but instead reflects a distributional mismatch between the noise conditions used during stage-2 fine-tuning and those encountered during prospective deployment~\cite{wotherspoon2021improved, sun2017unsupervised}. The noise signals used for stage-2 adaptation were collected from a single endoscopy center, representing center-specific equipment configurations and acoustic profiles. In contrast, prospective centers exhibited heterogeneous acoustic environments, reducing the direct applicability of learned noise patterns. Under these conditions, linguistic adaptation—captured in stage-1—contributed the majority of CER reduction, while additional noise adaptation yielded more subtle effects.

\textbf{Impact of content type and inter-center heterogeneity.}
As shown in Fig.~\ref{fig:prospective}b, EndoASR and EndoASR-noise achieve substantial improvements over the Paraformer baseline across all centers for lesion-related content. In contrast, for pre-procedural (PreOp) content, the performance differences between models remain limited.

This discrepancy can be attributed to the differing linguistic characteristics of these content types. Lesion-related descriptions contain a higher proportion of specialized medical terminology, whereas PreOp content is closer to general conversational language with lower terminology density. As the evaluation metric directly measures medical term accuracy, these results highlight the strong domain adaptation capability of the proposed models in terminology-intensive scenarios.

In addition, the pronounced improvement observed for IntraOp content in center 4 further reflects inter-center heterogeneity, suggesting that variations in clinical workflow and speech patterns across centers can influence model performance.
\textbf{Fine-grained semantic analysis of medical terminology.}
To further interpret the observed performance differences, we grouped medical terms into several high-level semantic categories based on their clinical meaning, including anatomical landmarks, lesion morphology, procedural actions, clinical context, quality descriptors, and lesion size expressions (Supplementary Fig.~S4).

This analysis shows that the largest improvements are concentrated in terminology-intensive categories such as lesion morphology and procedural expressions, where the proposed model achieves the highest accuracy gains over baseline systems. In contrast, anatomically grounded terms remain relatively stable across models, suggesting a smaller domain gap for these expressions.

Together, these findings demonstrate that the effectiveness of domain adaptation varies substantially across semantic categories, and provide a mechanistic explanation for the pronounced improvements observed in lesion-related content.

\textbf{Inference efficiency and real-time feasibility.}
Beyond recognition accuracy, inference latency is a critical factor for deployment in endoscopy units, where speech recognition must operate continuously and in near real time. As shown in Fig.~\ref{fig:F2}c, substantial differences in computational efficiency were observed across ASR models. The proposed domain-adapted model achieved one of the lowest RTF values while simultaneously delivering the strongest clinical accuracy. Although a small number of models exhibited marginally lower RTF, these gains came at the cost of substantially higher error rates (Fig.~\ref{fig:prospective}a). This trade-off underscores that, in high-acuity procedural environments, inference efficiency must be evaluated jointly with recognition reliability rather than in isolation.

\textbf{Limitations and future directions.}
This study has several limitations. First, noise recordings used for noise-aware adaptation were collected from a single center, which may limit robustness to unseen acoustic conditions. Second, linguistic adaptation relied primarily on synthesized speech derived from formal endoscopy reports, emphasizing professional terminology while underrepresenting spontaneous conversational patterns. Future work will incorporate multi-center noise data, accent-diverse clinician speech, and spontaneous intraoperative dialogue to further improve robustness and inclusivity.

Overall, our findings demonstrate that domain-adapted ASR can effectively bridge the out-of-distribution gap between general-purpose speech models and real-world endoscopy. With a moderate amount of synthetic data, substantial gains can be achieved, and these improvements generalize across heterogeneous multi-center settings, particularly in terminology-intensive clinical content. Notably, the proposed model achieves low real-time factor (RTF) with a compact parameter size (220M), enabling efficient and reliable deployment in real-time and edge scenarios.

These results position domain-adapted ASR as a practical and scalable solution for speech-based interaction in endoscopic clinical workflows.

\section{Method}\label{sec4}
\subsection{Dataset Construction}\label{sec4_1}
To support domain-specific automatic speech recognition (ASR) in endoscopic clinical scenarios, we constructed a comprehensive dataset comprising synthetic training data, retrospective clinical data, and prospective multi-center clinical data. Each component serves a distinct purpose in model training, robustness enhancement, and generalization evaluation.

\textbf{Clinical text corpus for endoscopic reporting.}
The textual source used for data construction was derived from authentic endoscopic examination reports routinely generated in clinical practice. These reports follow a highly structured and professional writing style, containing dense medical terminology and standardized descriptions of anatomical structures, pathological findings, and procedural outcomes.

Compared with conversational speech, endoscopic reports are characterized by a high proportion of domain-specific terms, long information-dense sentences, and strong structural regularity. Importantly, these reports encapsulate the terminology most frequently and accurately used by clinicians during endoscopic diagnosis and intervention, reflecting the core linguistic expressions employed in real procedural environments. This corpus provides high linguistic fidelity to real-world endoscopic documentation, which is essential for developing ASR systems tailored to clinical reporting tasks.

\textbf{Synthetic speech generation via text-to-speech.}
Due to the limited availability of large-scale annotated endoscopic speech data, a text-to-speech (TTS) strategy was adopted to generate synthetic speech aligned with the clinical text corpus~\cite{laptev2020you, ueno2021data, gokay2019improving}. All report texts were converted into Mandarin speech using a commercial-grade TTS system (\url{https://console.xfyun.cn/services/tts}
), with both male and female voices, ensuring consistent pronunciation of specialized medical terminology and precise audio–text alignment.

As shown in Fig.~\ref{fig:F2}a. Two types of synthetic audio were constructed: (1) clean synthetic speech generated directly from the TTS system, and (2) noise-augmented synthetic speech obtained by mixing clean synthetic speech with recorded background noise at controlled signal-to-noise ratios (SNRs). This design enables controlled investigation of domain-specific linguistic adaptation and acoustic robustness during model fine-tuning~\cite{li2014overview}.

\textbf{Operating-room noise collection and augmentation.}
To simulate realistic acoustic conditions encountered during endoscopic procedures, background noise was collected directly from endoscopy operating rooms during routine clinical practice. The recordings capture typical environmental sounds, including endoscopy tower operation, flushing and suction devices, alarms, human movement, and multi-speaker interactions.

All noise samples were recorded at a single clinical center, reflecting authentic but center-specific acoustic characteristics. Noise recordings were obtained across multiple functional environments within the endoscopy unit, including examination rooms, interventional procedure areas, and patient preparation spaces, thereby capturing diverse yet institution-specific noise profiles. To characterize real-world acoustic conditions, noise statistics were quantified using intraoperative recordings from a retrospective clinical dataset (described below), in which the SNR ranged from 23--40~dB (mean: 29.81~dB, SD: 3.62~dB).

Guided by these empirical measurements, noise-augmented synthetic speech was generated using a deliberately more challenging SNR distribution of 20--28~dB (mean: 23.98~dB, SD: 1.16~dB). This conservative design encourages robustness under realistic and adverse operating-room conditions, facilitating effective adaptation of ASR models to real-world endoscopy environments.

\textbf{Retrospective clinical speech dataset.}
The retrospective dataset comprised 600 real intraoperative speech segments collected from a single clinical center during routine endoscopic practice. Six endoscopists participated in data collection, and each endoscopist contributed 100 utterance-level segments. These samples represent naturally occurring endoscopic speech and were recorded under acoustic conditions similar to those of the endoscopy-room noise recordings used for synthetic data augmentation. This dataset was primarily used for retrospective benchmarking under realistic but center-specific conditions.

\textbf{Prospective multi-center clinical speech dataset. }
To evaluate model generalization across institutions, we prospectively collected a multi-center dataset consisting of 300 real intraoperative speech segments from five independent endoscopy centers, with each center contributing 60 unique audio–text pairs. The prospective data had an average audio duration of approximately 10 seconds, an average transcript length of 40.4 characters, and an average SNR of 31.53 dB (range: 23–48 dB). Within each center, the 60 samples were evenly distributed across six predefined clinical content categories, with 10 samples per category: (A) PreOp, prior medical history and indication for the current examination; (B) IntraOp, bowel preparation quality and intra-procedural patient status; (C) Lesion, colorectal polyp lesions and corresponding interventions; (D) Cancer, colorectal malignant tumors and corresponding management; (E) Inflammation, inflammatory bowel disease and related descriptive findings; and (F) PostOp, examination conclusions and post-procedural follow-up plans. This design enabled fine-grained evaluation across centers, disease types, and procedural phases under real-world deployment conditions.

\textbf{Gold-standard transcription and verification. }
For both the retrospective and prospective datasets, each speech segment was paired with a gold-standard reference transcript through a clinician-verified annotation workflow. A verbatim transcript was first prepared from the raw audio, after which the speaking endoscopist replayed the segment and manually checked the transcript against the original utterance, with particular attention to medical terminology, anatomical locations, procedural actions, and numeric expressions. Any ambiguous or inconsistent content was rechecked directly against the audio until the written transcript was fully consistent with the spoken content. The finalized endoscopists-confirmed transcripts were used as the reference standard for all quantitative evaluations.

\textbf{Transcription rulebook. }
Transcripts were prepared in verbatim Mandarin Chinese and preserved the speaker’s final intended clinical wording without paraphrasing. Standardized endoscopic terms, anatomical sites, lesion classifications, procedure names, drug names, and explicitly spoken scores or numbers were recorded in a consistent written format. Brief fillers, hesitations, and aborted self-repairs that did not contribute to the final clinical meaning were omitted, whereas repeated clinically meaningful content was retained. Punctuation was used only to mark clear utterance boundaries and was not used to infer information not present in the audio.

\subsection{Model Architecture and Training Strategy}\label{sec4_2}
\textbf{Base ASR model.}
We adopt Seaco-Paraformer(\url{https://www.modelscope.cn/models/iic/speech_seaco_paraformer_large_asr_nat-zh-cn-16k-common-vocab8404-pytorch/files}) as the base automatic speech recognition (ASR) architecture in this study. Seaco-Paraformer is a non-autoregressive end-to-end ASR model that decouples acoustic modeling and linguistic prediction via a predictor--encoder--decoder framework, enabling efficient inference while maintaining competitive transcription accuracy.

The model is initialized with publicly available pretrained weights trained on large-scale general-domain Mandarin speech corpora. No architectural modifications are introduced, allowing the investigation to focus on data-driven domain adaptation rather than model redesign.

\textbf{Two-stage domain adaptation strategy.}
To effectively adapt the pretrained ASR model to the endoscopic clinical domain, we employ a two-stage fine-tuning strategy. This design explicitly separates linguistic domain adaptation from acoustic robustness learning, reducing optimization interference between heterogeneous data sources~\cite{gao2024enhancing}.

In the first stage (EndoASR), the model is fine-tuned exclusively using clean synthetic speech generated from the clinical text corpus. These data are acoustically clean and precisely aligned, enabling the model to adapt its language modeling capacity toward domain-specific terminology, sentence structures, and reporting conventions without the confounding effect of background noise.

In the second stage (EndoASR-noise), the EndoASR model is further fine-tuned using noise-augmented synthetic speech. The augmented data are constructed by mixing the same synthetic speech with operating-room noise at varying signal-to-noise ratios. This stage focuses on improving robustness to realistic acoustic disturbances encountered during endoscopic procedures while preserving the domain-specific linguistic knowledge learned in the first stage.

\textbf{Training objective and optimization.}
All fine-tuning stages use the standard Seaco-Paraformer training objective, combining the attention-based decoder loss and the predictor-related auxiliary losses as defined in the original implementation. The character error rate (CER) is not directly optimized but is used as the primary evaluation metric.

Training is conducted using the Adam optimizer with a fixed learning rate schedule consistent across stages. Early stopping is applied based on validation loss to prevent overfitting. All hyperparameters are kept identical between EndoASR and EndoASR-noise to ensure fair comparison and controlled analysis.

\subsection{Evaluation Protocol and Metrics}\label{sec4_3}
\textbf{Evaluation datasets.}
Model performance is evaluated on three distinct datasets to comprehensively assess domain adaptation and real-world generalization: (1) a synthetic test set, (2) a retrospective real-world dataset, and (3) a prospective multi-center real-world dataset.

The synthetic test set consists of held-out synthetic speech samples constructed from clinical reports, including both clean and noise-augmented audio, and is used to assess domain-specific language adaptation and noise robustness under controlled conditions.

The retrospective real-world dataset comprises intraoperative speech recordings collected from a single center, involving six endoscopists. These data are used to evaluate model performance under real clinical conditions with speaker variability while maintaining consistent acoustic environments.

The prospective dataset includes recordings collected across five independent endoscopy centers during routine clinical practice. This dataset reflects real deployment scenarios, incorporating inter-center variability in acoustic environments, workflow patterns, and speaking styles. No data from these centers are used during model training or adaptation.

\textbf{Evaluation protocol.}
All models are evaluated in an offline transcription setting. For retrospective and prospective evaluations, each audio recording is paired with a manually verified reference transcription.

To enable speaker-level and center-level analysis, metrics are computed separately for each endoscopist in the retrospective dataset and for each center in the prospective dataset, in addition to overall averages.

\textbf{Evaluation metrics.}
We report complementary metrics to evaluate transcription quality from lexical, semantic, and clinically relevant perspectives:
\begin{itemize}
\item \textbf{Character Error Rate (CER)}: defined as the normalized Levenshtein distance between the predicted and reference character sequences, serving as a coarse-grained accuracy metric for Mandarin ASR.
\item \textbf{BLEU-1}: computed at the unigram level to measure surface-level lexical overlap between predictions and references~\cite{papineni2002bleu}.
\item \textbf{BERTScore}: calculated using contextualized embeddings to quantify semantic similarity between predicted and reference texts~\cite{zhang2019bertscore}.
\item \textbf{Medical terminology accuracy (Med ACC)}: a domain-specific metric that measures the proportion of correctly recognized medical terms within each utterance. The terminology list was curated by endoscopists from the prospective dataset and includes clinically critical entities such as lesion types, anatomical locations, procedural actions, and standardized classifications (e.g., BBPS, Paris classification). A term is considered correct only if all constituent characters are accurately transcribed in sequence. This metric directly reflects the reliability of ASR systems for clinical documentation and downstream decision support.
\end{itemize}
All metrics are reported as mean $\pm$ standard deviation across samples unless otherwise specified.

\textbf{Signal-to-noise ratio analysis.}
For analyses involving acoustic robustness, the wideband signal-to-noise ratio (SNR) of each real-world recording was estimated using a WADA-SNR--style amplitude statistics--based approach inspired by the WADA framework~\cite{kim2008robust}. This method derives a reference-free proxy of relative noise conditions from the distributional properties of the speech waveform and is suitable for large-scale, unconstrained clinical recordings where ground-truth noise references are unavailable. Importantly, SNR estimates were used exclusively for stratified analysis and correlation studies and were not utilized during model training or adaptation.

\subsection{Implementation Details}
Model fine-tuning was performed using the FunASR~\cite{gao2023funasr} training framework. All experiments followed a unified training configuration across both adaptation stages. Token-based batching was adopted to accommodate variable-length clinical speech, with a maximum batch size of 6000 tokens and a sorting buffer size of 1024. The batch type was set to token-level batching, and data loading was parallelized with four workers.
Training was conducted for up to 50 epochs using the Adam optimizer with an initial learning rate of $2 \times 10^{-4}$. Model checkpoints were saved and validated every 2000 training steps. The top 20 checkpoints ranked by validation loss were retained, and the final model for each stage was obtained by averaging the best 10 checkpoints. 
All training and validation were performed on an NVIDIA RTX 5090. No additional architectural modifications, external language models, or post-processing heuristics were introduced during training or inference.

\section{Data availability}\label{sec5}
The synthetic training dataset generated from structured endoscopy reports will be publicly released upon publication.
De-identified subsets of the retrospective and prospective clinical datasets will also be made available to the research community, subject to institutional approval and data use agreements, to facilitate reproducibility and further research.
All remaining data are available from the corresponding authors upon reasonable request.
\section{Code availability}\label{sec6}
The code is available at \url{https://github.com/ku262/EndoASR}

\backmatter

\bibliography{main}

\end{document}